\documentclass[sigconf]{acmart}

\AtBeginDocument{%
  \providecommand\BibTeX{{%
    \normalfont B\kern-0.5em{\scshape i\kern-0.25em b}\kern-0.8em\TeX}}}

\copyrightyear{2022} 
\acmYear{2022} 
\setcopyright{acmcopyright}\acmConference[ICSE '22 Companion]{44th International Conference on Software Engineering Companion}{May 21--29, 2022}{Pittsburgh, PA, USA}
\acmBooktitle{44th International Conference on Software Engineering Companion (ICSE '22 Companion), May 21--29, 2022, Pittsburgh, PA, USA}
\acmPrice{15.00}
\acmDOI{10.1145/3510454.3516865}
\acmISBN{978-1-4503-9223-5/22/05}

\usepackage{subcaption}
\usepackage{caption}
\usepackage{multirow}
\usepackage{enumitem}
\usepackage{arydshln}

\begin{document}

\title{ReGVD: Revisiting Graph Neural Networks for Vulnerability Detection}

\author{Van-Anh Nguyen}
\affiliation{\institution{VNU - University of Science, Vietnam}}
\email{vananhnt57@gmail.com}
\authornote{The first two authors contributed equally to this work.}

\author{Dai Quoc Nguyen}
\affiliation{\institution{Oracle Labs, Australia}}
\email{dai.nguyen@oracle.com}
\authornotemark[1]
\authornote{Corresponding author.
This work was done before Dai Quoc Nguyen joined Oracle Labs, Australia.}

\author{Van Nguyen}
\affiliation{\institution{Monash University, Australia}}
\email{van.nk@monash.edu}

\author{Trung Le}
\affiliation{\institution{Monash University, Australia}}
\email{trunglm@monash.edu}

\author{Quan Hung Tran}
\affiliation{\institution{Adobe Research, San Jose, CA, USA}}
\email{qtran@adobe.com}

\author{Dinh Phung}
\affiliation{\institution{Monash University, Australia}}
\email{dinh.phung@monash.edu}



\begin{abstract}
Identifying vulnerabilities in the source code is essential to protect the software systems from cyber security attacks. It, however, is also a challenging step that requires specialized expertise in security and code representation. To this end, we aim to develop a general, practical, and programming language-independent model capable of running on various source codes and libraries without difficulty. Therefore, we consider vulnerability detection as an inductive text classification problem and propose ReGVD, a simple yet effective graph neural network-based model for the problem. In particular, ReGVD views each raw source code as a flat sequence of tokens to build a graph, wherein node features are initialized by only the token embedding layer of a pre-trained programming language (PL) model. ReGVD then leverages residual connection among GNN layers and examines a mixture of graph-level sum and max poolings to return a graph embedding for the source code. ReGVD outperforms the existing state-of-the-art models and obtains the highest accuracy on the real-world benchmark dataset from CodeXGLUE for vulnerability detection. Our code is available at: \url{https://github.com/daiquocnguyen/GNN-ReGVD}.

\end{abstract}

\maketitle

\section{Introduction}
The software vulnerability problems have rapidly grown recently, either reported through publicly disclosed information-security flaws and exposures (CVE) or exposed inside privately-owned source codes and open-source libraries. 
These vulnerabilities are the main reasons for cyber security attacks on the software systems that cause substantial damages economically and socially \citep{neuhaus2007predicting,devign2019}. 
Therefore, vulnerability detection is an essential yet challenging step to identify vulnerabilities in the source codes to provide security solutions for the software systems. 

Early approaches \citep{neuhaus2007predicting,nguyen2010predicting,shin2010evaluating} have been proposed to carefully design hand-engineered features for machine learning algorithms to detect vulnerabilities. These early approaches, however, suffer from two major drawbacks. First, creating good features requires prior knowledge, hence needs domain experts, and is usually time-consuming. Second, hand-engineered features are impractical and not straightforward to adapt to all vulnerabilities in numerous open-source codes and libraries evolving over time.

To reduce human efforts on feature engineering, some approaches \citep{li2018vuldeepecker,russell2018automated} consider each raw source code as a flat natural language sequence and explore deep learning architectures applied for natural language processing (NLP) (such as LSTMs \citep{lstm1997} and CNNs \citep{textcnn2014}) in detecting vulnerabilities. 
Recently, pre-trained language models such as BERT \citep{devlin2018bert} have emerged as a trending learning paradigm, achieving significant success in NLP applications. 
Inspired by this BERT-style trending paradigm, pre-trained programming language (PL) models such as CodeBERT \citep{codebert2020} have improved the performance of PL downstream tasks such as vulnerability detection. 
However, 
as mentioned in \citep{Nguyen2019UGT}, 
all interactions among all positions in the input sequence inside the self-attention layer of the BERT-style model build up a complete graph, i.e., every position has an edge to all other positions; thus, this limits learning local structures within the source code to differentiate vulnerabilities.

\begin{figure*}[!ht]
\centering
\includegraphics[width=0.75\textwidth]{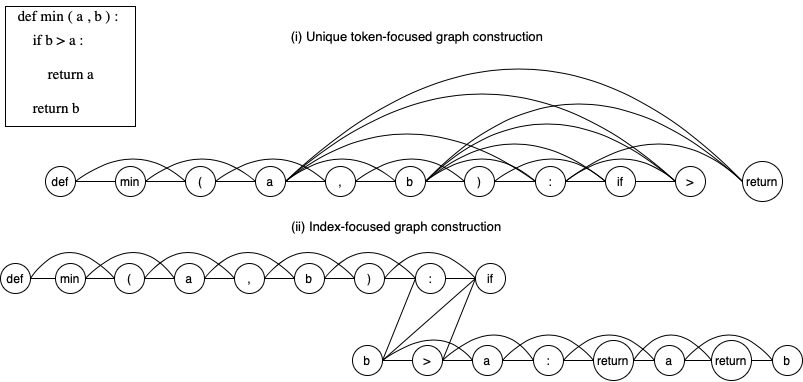}
\captionof{figure}{An illustration for two graph construction methods with a fixed-size sliding window of length 3.}
\label{fig:graphconstruction}
\end{figure*}

Graph neural networks (GNNs) have recently become a primary method to embed nodes and graphs into low-dimensional continuous vector spaces \citep{hamilton2017representation,wu2019comprehensive,NGUYEN2021Thesis}.
GNNs provide faster and practical training, higher accuracy, and state-of-the-art results for downstream tasks such as text classification \citep{yao2019graph,huang-etal-2019-text,zhang-etal-2020-every,Nguyen2021QGNN}. 
Devign \citep{devign2019} is proposed to utilize Gated GNNs \citep{li2015gated} for vulnerability detection, wherein 
Devign uses a PL parser to extract multi-edged graph information.
However, Devign is difficult of being practiced in reality.
The main reason is that there is not a perfect parser in reality for each PL, which can successfully parse a variety of source codes and libraries without any internal compile errors and exceptions.

In this paper, our goal is to develop a general, practical, and programming language-independent model capable of running on various source codes and libraries without difficulty.
Hence, we consider vulnerability detection as an inductive text classification problem and introduce ReGVD -- a simple yet effective GNN-based model for vulnerability detection as follows: (i) ReGVD views each raw source code as a flat sequence of tokens to construct a graph (in Section \ref{subsec:graphconstruction}), wherein node features are initialized by only the token embedding layer of a pre-trained PL model.
(ii) ReGVD leverages GNNs (such as GCNs \citep{kipf2017semi} or Gated GNNs \citep{li2015gated}) using residual connection among GNN layers (in Section \ref{subsec:rgnn}).
(iii) ReGVD examines a mixture between the sum and max poolings to produce a graph embedding for the source code (in Section \ref{subsec:readout}). 
This graph embedding is fed to a single fully-connected layer followed by a softmax layer to predict the code vulnerabilities.
Extensive experiments show that ReGVD significantly outperforms the existing state-of-the-art models on the benchmark vulnerability detection dataset from CodeXGLUE \citep{CodeXGLUE2021}. ReGVD produces the highest accuracy of 63.69\%, gaining absolute improvements of 1.61\% and 1.39\% over CodeBERT and GraphCodeBERT, respectively; thus, ReGVD can act as a new strong baseline for future work.

\section{The proposed ReGVD}

\subsection{Problem definition}
\label{sec:probdef}

We consider vulnerability detection for source code at the function level, i.e., we aim to identify whether a given function in raw source code is vulnerable or not \citep{devign2019}.
We define a data sample as $\left\{(\mathsf{c}_i, \mathsf{y}_i) | \mathsf{c}_i \in \mathbb{C}, \mathsf{y}_i \in \mathbb{Y}\right\}_{i=1}^n$, where $\mathbb{C}$ represents the set of raw source codes, $\mathbb{Y}=\{0,1\}$ denotes the label set with $1$ for vulnerable and $0$ otherwise, and $n$ is the number of instances. 
In this work, we consider vulnerability detection as an inductive text classification problem and leverage GNNs for the problem.
Therefore, we construct a graph $\mathsf{g}_i(\mathcal{V}, \boldsymbol{X}, \boldsymbol{A}) \in \mathcal{G}$ for each source code $\mathsf{c}_i$, wherein $\mathcal{V}$ is a set of $m$ nodes in the graph; $\boldsymbol{X} \in \mathbb{R}^{m \times d}$ is the node feature matrix, wherein each node $\mathsf{v}_j \in \mathcal{V}$ is represented by a $d$-dimensional real-valued vector $\boldsymbol{x}_j \in \mathbb{R}^{d}$; $\boldsymbol{A} \in \{0, 1\}^{m \times m}$ is the adjacency matrix, where $\boldsymbol{A}_{\mathsf{v},\mathsf{u}}$ equal to 1 means having an edge between node $\mathsf{v}$ and node $\mathsf{u}$, and 0 otherwise. 
We aim to learn a mapping function $f: \mathcal{G} \rightarrow \mathbb{Y}$ to determine whether a given source code is vulnerable or not. 
The mapping function $f$ can be learned by minimizing the loss function with the regularization on model parameters $\boldsymbol{\theta}$ as:
\begin{equation}
\mathsf{min} \sum_{i=1}^{n} \mathcal{L}(f(\mathsf{g}_i(\mathcal{V}, \boldsymbol{X}, \boldsymbol{A}), \mathsf{y}_i|\mathsf{c}_i)) +
\lambda\|\boldsymbol{\theta}\|_2^2 \nonumber
\end{equation}
where $\mathcal{L}(.)$ is the cross-entropy loss function and and $\lambda$ is an adjustable weight.



\subsection{Graph construction}
\label{subsec:graphconstruction}

We consider a raw source code as a flat sequence of tokens and illustrate two graph construction methods \citep{huang-etal-2019-text,zhang-etal-2020-every} in Figure \ref{fig:graphconstruction}, wherein we omit self-loops in these two methods since the self-loops do not help to improve performance in our pilot experiments.\footnote{In our implementation, we firstly tokenize the source code using the corresponding tokenizer of the pre-trained PL model, and then we construct the graph from the tokenized sequence.}

\paragraph{Unique token-focused construction} We represent unique tokens as nodes and co-occurrences between tokens (within a fixed-size sliding window) as edges, and the obtained graph has an adjacency matrix $\boldsymbol{A}$ as:
\begin{equation}
    \boldsymbol{A}_{\mathsf{v},\mathsf{u}} = \left\{ 
        \begin{array}{l}
            1 \ \ \ \text{If $\mathsf{v}$ and $\mathsf{u}$ co-occur within a sliding window} \\  \ \ \ \ \ \ \ \ \ \ \text{and $\mathsf{v} \neq \mathsf{u}$.}\\
            0 \ \ \ \text{Otherwise.}
        \end{array} \right. \nonumber
\end{equation}

\paragraph{Index-focused construction} Given a flat sequence of $l$ tokens $\left\{t_\mathsf{i}\right\}_{\mathsf{i}=1}^l$, we represent all tokens as the nodes, i.e., treating each index $\mathsf{i}$ as a node to represent token $t_\mathsf{i}$. The number of nodes equals the sequence length.
We also consider co-occurrences between indexes (within a fixed-size sliding window) as edges, and the obtained graph has an adjacency matrix $\boldsymbol{A}$ as:
\begin{equation}
    \boldsymbol{A}_{\mathsf{i},\mathsf{j}} = \left\{ 
        \begin{array}{l}
            1 \ \ \ \text{If $\mathsf{i}$ and $\mathsf{j}$ co-occur within a sliding window}\\ \ \ \ \ \ \ \ \ \ \ \text{and $\mathsf{i} \neq \mathsf{j}$.}\\
            0 \ \ \ \text{Otherwise.}
        \end{array} \right. \nonumber
\end{equation}

\paragraph{Node feature initialization} It is worth noting that pre-trained programming language (PL) models such as CodeBERT \citep{codebert2020} have recently improved the performance of PL downstream tasks such as vulnerability detection.
To attain the advantage of the pre-trained PL model and also to make \textit{a fair comparison}, we use \textit{only the token embedding layer} of the pre-trained PL model to initialize node feature vectors for reporting our final results.


\subsection{Graph neural networks with residual connection}
\label{subsec:rgnn}

GNNs aim to update vector representations of nodes by recursively aggregating vector representations from their neighbours \citep{scarselli2009graph,kipf2017semi}.
Mathematically, given a graph $\mathsf{g}(\mathcal{V}, \boldsymbol{X}, \boldsymbol{A})$, we simply formulate GNNs as follows:
\begin{eqnarray}
\boldsymbol{\mathsf{H}}^{(k+1)} = \mathsf{GNN}\left(\boldsymbol{A}, \boldsymbol{\mathsf{H}}^{(k)}\right) \nonumber
\end{eqnarray}
where $\boldsymbol{\mathsf{H}}^{(k)}$ is the matrix representation of nodes at the $k$-th iteration/layer; 
and $\boldsymbol{\mathsf{H}}^{(0)} = \boldsymbol{X}$.
There have been many GNNs proposed in recent literature \citep{wu2019comprehensive}, wherein Graph Convolutional Networks (GCNs) \citep{kipf2017semi} is the most widely-used one, and Gated graph neural networks (``Gated GNNs'' or ``GGNNs'' for short) \citep{li2015gated} is also suitable for our data structure. Our ReGVD leverages GCNs and GGNNs as the base models.

Formally, GCNs is given as follows:
\begin{equation}
\boldsymbol{\mathsf{h}}_{\mathsf{v}}^{(k+1)} = \phi\left(\sum_{\mathsf{u} \in \mathcal{N}_\mathsf{v}}a_{\mathsf{v},\mathsf{u}}\boldsymbol{W}^{(k)}\boldsymbol{\mathsf{h}}_{\mathsf{u}}^{(k)}\right) , \forall \mathsf{v} \in \mathcal{V} \nonumber
\label{equa:gcn}
\end{equation}
where $a_{\mathsf{v},\mathsf{u}}$ is an edge constant between nodes $\mathsf{v}$ and $\mathsf{u}$ in the Laplacian re-normalized adjacency matrix $\textbf{D}^{-\frac{1}{2}}\boldsymbol{A}\textbf{D}^{-\frac{1}{2}}$ (as we omit self-loops), wherein $\textbf{D}$ is the diagonal node degree matrix of $\boldsymbol{A}$; $\boldsymbol{W}^{(k)}$ is a weight matrix; and $\phi$ is a nonlinear activation function such as $\mathsf{ReLU}$.

GGNNs adopts GRUs \citep{cho2014learning}, unrolls the recurrence for a fixed number of timesteps, and removes the need to constrain parameters to ensure convergence as:
\begin{eqnarray}
\textbf{a}_{\mathsf{v}}^{(k+1)} &=& \sum_{\mathsf{u} \in \mathcal{N}_\mathsf{v}}a_{\mathsf{v},\mathsf{u}}\boldsymbol{\mathsf{h}}_{\mathsf{u}}^{(k)} \nonumber\\
\textbf{z}_{\mathsf{v}}^{(k+1)} &=& \sigma\left(\textbf{W}^{z}\textbf{a}_{\mathsf{v}}^{(k+1)} + \textbf{U}^{z}\boldsymbol{\mathsf{h}}_{\mathsf{v}}^{(k)}\right) \nonumber \\
\textbf{r}_{\mathsf{v}}^{(k+1)} &=& \sigma\left(\textbf{W}^{r}\textbf{a}_{\mathsf{v}}^{(k+1)} + \textbf{U}^{r}\boldsymbol{\mathsf{h}}_{\mathsf{v}}^{(k)}\right) \nonumber \\
\widetilde{\boldsymbol{\mathsf{h}}_{\mathsf{v}}^{(k+1)}} &=& \phi\left(\textbf{W}^{o}\textbf{a}_{\mathsf{v}}^{(k+1)} + \textbf{U}^{o}\left(\textbf{r}_{\mathsf{v}}^{(k+1)}\odot\boldsymbol{\mathsf{h}}_{\mathsf{v}}^{(k)}\right)\right) \nonumber \\
\boldsymbol{\mathsf{h}}_{\mathsf{v}}^{(k+1)} &=& \left(1 - \textbf{z}_{\mathsf{v}}^{(k+1)}\right)\odot\boldsymbol{\mathsf{h}}_{\mathsf{v}}^{(k)} + \textbf{z}_{\mathsf{v}}^{(k+1)}\odot\widetilde{\boldsymbol{\mathsf{h}}_{\mathsf{v}}^{(k+1)}} \nonumber \label{equa:gatedgnn}
\end{eqnarray}
where $\textbf{z}$ and $\textbf{r}$ are the update and reset gates; $\sigma$ is the sigmoid function; and $\odot$ is the element-wise multiplication.

The residual connection \citep{he2016deep} is used to incorporate information learned in the lower layers to the higher layers, and more importantly, to allow gradients to directly pass through the layers to avoid vanishing gradient or exploding gradient problems.
Motivated by that, we follow \citep{bresson2017residual} to adapt residual connection among the GNN layers, with fixing the same hidden size for the different layers.
In particular, ReGVD redefines GNNs as:
\begin{equation}
\boldsymbol{\mathsf{H}}^{(k+1)} = \boldsymbol{\mathsf{H}}^{(k)} + \mathsf{GNN}\left(\boldsymbol{A}, \boldsymbol{\mathsf{H}}^{(k)}\right) \nonumber
\end{equation}


\begin{figure}[!ht]
\centering
\includegraphics[width=0.5\textwidth]{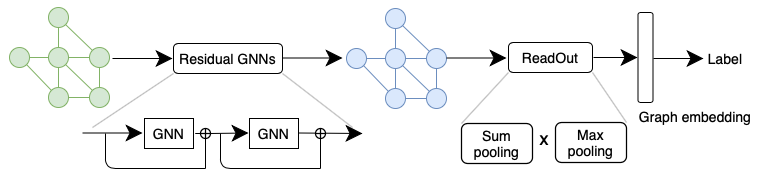}
\captionof{figure}{An illustration for our proposed ReGVD.}
\label{fig:regvd}
\end{figure}

\subsection{Graph-level readout pooling layer}
\label{subsec:readout}

The graph-level readout layer is used to produce a graph embedding for each input graph.   
ReGVD leverages the sum pooling as it produces better results for graph classification \citep{xu2019powerful}.\footnote{In our pilot studies, using the sum pooling $\sum_{\mathsf{v} \in \mathcal{V}}\boldsymbol{\mathsf{e}}_\mathsf{v}$ also provides higher accuracies than using the mean pooling $\frac{1}{|\mathcal{V}|}\sum_{\mathsf{v} \in \mathcal{V}}\boldsymbol{\mathsf{e}}_\mathsf{v}$ employed in \citep{zhang-etal-2020-every}.}
Besides, ReGVD utilizes the max pooling to exploit more information on the node representations.
ReGVD then considers a mixture between the sum and max poolings to produce the graph embedding $\boldsymbol{\mathsf{e}}_{\mathsf{g}}$ as:
\begin{eqnarray}
\boldsymbol{\mathsf{e}}_\mathsf{v} &=& \sigma\left(\textbf{w}^\mathsf{T}\boldsymbol{\mathsf{h}}_\mathsf{v}^{(K)} + \mathsf{b}\right)\odot\phi\left(\textbf{W}\boldsymbol{\mathsf{h}}_{\mathsf{v}}^{(K)} + \boldsymbol{\mathsf{b}}\right) \nonumber \\ 
\boldsymbol{\mathsf{e}}_{\mathsf{g}} &=& \mathsf{MIX}\left(\sum_{\mathsf{v} \in \mathcal{V}}\boldsymbol{\mathsf{e}}_\mathsf{v}, \textsc{MaxPool}\left\{\boldsymbol{\mathsf{e}}_\mathsf{v}\right\}_{\mathsf{v} \in \mathcal{V}}\right) \nonumber
\end{eqnarray}
where $\boldsymbol{\mathsf{e}}_\mathsf{v}$ is the final vector representation of node $\mathsf{v}$, wherein $\sigma\left(\textbf{w}^\mathsf{T}\boldsymbol{\mathsf{h}}_\mathsf{v}^{(K)} + \mathsf{b}\right)$ acts as soft attention mechanisms over nodes \citep{li2015gated}, and $\boldsymbol{\mathsf{h}}_\mathsf{v}^{(K)}$ is the vector representation of node $\mathsf{v}$ at the last $K$-th layer; and $\mathsf{MIX}(.)$ denotes an arbitrary function.
ReGVD examines three $\mathsf{MIX}$ functions consisting of $\mathsf{SUM}$, $\mathsf{MUL}$, and $\mathsf{CONCAT}$ as:
\begin{eqnarray}
\mathsf{SUM}&:& \boldsymbol{\mathsf{e}}_{\mathsf{g}} = \sum_{\mathsf{v} \in \mathcal{V}}\boldsymbol{\mathsf{e}}_\mathsf{v} +  \textsc{MaxPool}\left\{\boldsymbol{\mathsf{e}}_\mathsf{v}\right\}_{\mathsf{v} \in \mathcal{V}} \nonumber \\
\mathsf{MUL}&:& \boldsymbol{\mathsf{e}}_{\mathsf{g}} = \sum_{\mathsf{v} \in \mathcal{V}}\boldsymbol{\mathsf{e}}_\mathsf{v} \odot \textsc{MaxPool}\left\{\boldsymbol{\mathsf{e}}_\mathsf{v}\right\}_{\mathsf{v} \in \mathcal{V}} \nonumber \\
\mathsf{CONCAT}&:& \boldsymbol{\mathsf{e}}_{\mathsf{g}} = \left[\sum_{\mathsf{v} \in \mathcal{V}}\boldsymbol{\mathsf{e}}_\mathsf{v} \ \ \ \| \ \ \ \textsc{MaxPool}\left\{\boldsymbol{\mathsf{e}}_\mathsf{v}\right\}_{\mathsf{v} \in \mathcal{V}}\right] \nonumber
\end{eqnarray}
After that, ReGVD feeds $\boldsymbol{\mathsf{e}}_{\mathsf{g}}$ to a single fully-connected layer followed by a $\mathsf{softmax}$ layer to predict whether the source code is vulnerable or not as:
$\boldsymbol{\mathsf{\hat{y}}}_{\mathsf{g}} = \mathsf{softmax}\left(\textbf{W}_1\boldsymbol{\mathsf{e}}_{\mathsf{g}} + \textbf{b}_1\right)$
Finally, ReGVD is trained by minimizing the cross-entropy loss function as mentioned in Section \ref{sec:probdef}. 
We illustrate the proposed ReGVD in Figure \ref{fig:regvd}.












\section{Experimental setup and results}  \label{sec:experiment}





\subsection{Experimental setup}


\paragraph{Dataset} 
We use the real-world benchmark from CodeXGLUE \citep{CodeXGLUE2021} for vulnerability detection at the function level.\footnote{\url{https://github.com/microsoft/CodeXGLUE/tree/main/Code-Code/Defect-detection}}
The dataset was firstly created by \citet{devign2019}, including 27,318 manually-labeled vulnerable or non-vulnerable functions extracted from security-related commits in two large and popular C programming language open-source projects (i.e., QEMU and FFmpeg) and diversified in functionality.
Then \citet{CodeXGLUE2021} combined these projects and then split into the training/validation/test sets. 

\paragraph{Training protocol} 
We construct a 2-layer model, set the batch size to 128, and employ the Adam optimizer \citep{kingma2014adam} to train our model up to 100 epochs.
As mentioned in Section \ref{subsec:rgnn}, we set the same hidden size (``hs'') for the hidden GNN layers, wherein we vary the size value in \{128, 256, 384\}.
We vary the sliding window size (``ws'') in \{2, 3, 4, 5\} and the Adam initial learning rate (``lr'') in $\left\{1e^{-4}, 5e^{-4}, 1e^{-3}\right\}$.
The final accuracy on the test set is reported for the best model checkpoint, which obtains the highest accuracy on the validation set. 


\paragraph{Baselines}

We compare our ReGVD with strong and up-to-date baselines 
as follows:
\begin{itemize}

\item \textbf{BiLSTM} \citep{lstm1997} and \textbf{TextCNN} \citep{textcnn2014} are two well-known standard models applied for text classification. 

\item \textbf{RoBERTa} \citep{liu2019roberta} is built based on BERT \citep{devlin2018bert} by removing the next-sentence objective and training on a massive dataset with larger mini-batches and learning rates.

\item \textbf{Devign} \citep{devign2019} builds a multi-edged graph from a raw source code, then uses Gated GNNs \citep{li2015gated} to update node representations, and finally utilizes a 1-D CNN-based pooling (``\textit{Conv}'') to make a prediction. 
We note that \citet{devign2019} did not release the official implementation of Devign. 
Thus, we re-implement Devign using the same training and evaluation protocols.

\item \textbf{CodeBERT} \citep{codebert2020} is a pre-trained model also based on BERT for 6 programming languages (Python, Java, JavaScript, PHP, Ruby, Go), using masked language model \citep{devlin2018bert} and replaced token detection \citep{clark2020electra} objectives.

\item \textbf{GraphCodeBERT} \citep{graphcodebert2021} is a new pre-trained PL model, extending CodeBERT to consider the inherent structure of code data flow into the training objective.

\end{itemize}


\subsection{Main results}

\begin{table}[!ht]
\centering
\caption{Vulnerability detection accuracies (\%) on the test set.
The best scores are in {bold}, while the second best scores are in underline. 
The results of BiLSTM, TextCNN, RoBERTa, and CodeBERT are taken from \citep{CodeXGLUE2021}. 
$\star$ denotes that we report our own results for other baselines.
``Idx'' and ``UniT'' denote the index-focused graph construction and the unique token-focused one, respectively. ``CB'' and ``G-CB'' denote using only {the token embedding layer} of CodeBERT and GraphCodeBERT to initialize the node features, respectively.
}
\def\arraystretch{1.1}
\begin{tabular}{lc}
\multirow{1}{*}{\textbf{Model}} & \textbf{Accuracy}\\
\hline 
BiLSTM & 59.37\\
TextCNN & 60.69\\
RoBERTa & 61.05\\
CodeBERT & 62.08\\
\hline
GraphCodeBERT$^\star$ & 62.30\\
\hdashline
Devign (Idx + CB)$^\star$ & 60.43\\
Devign (Idx + G-CB)$^\star$ & 61.31\\
Devign (UniT + CB)$^\star$ & 60.40\\
Devign (UniT + G-CB)$^\star$ & 59.77\\
\hline 
\textbf{ReGVD} (GGNN + Idx + CB) & 63.54\\
\textbf{ReGVD} (GGNN + Idx + G-CB) & 63.29\\
\textbf{ReGVD} (GGNN + UniT + CB) & \underline{63.62}\\
\textbf{ReGVD} (GGNN + UniT + G-CB) & 62.41\\
\hdashline
\textbf{ReGVD} (GCN + Idx + CB) & 62.63\\
\textbf{ReGVD} (GCN + Idx + G-CB) & 62.70\\
\textbf{ReGVD} (GCN + UniT + CB) & 63.14\\
\textbf{ReGVD} (GCN + UniT + G-CB) & \textbf{63.69}\\
\hline
\end{tabular}
\label{table:result}
\end{table}

Table \ref{table:result} presents the accuracy results of the proposed ReGVD and the strong and up-to-date baselines on the real-world benchmark dataset from CodeXGLUE for vulnerability detection.
We note that 
both the recent models CodeBERT and GraphCodeBERT obtain competitive performances and perform better than Devign, indicating the effectiveness of the pre-trained PL models.
More importantly, ReGVD gains absolute improvements of 1.61\% and 1.39\% over CodeBERT and GraphCodeBERT, respectively. 
This shows the benefit of ReGVD in learning the local structures inside the source code to differentiate vulnerabilities (w.r.t using only the token embedding layer of the pre-trained PL model).
Hence, our ReGVD significantly outperforms the up-to-date baseline models. 
In particular, ReGVD produces the highest accuracy of 63.69\% -- a new state-of-the-art result on the CodeXGLUE vulnerability detection dataset.

\begin{figure}[!ht]
\centering
\begin{subfigure}{.245\textwidth}
  \centering
  \includegraphics[width=1\textwidth]{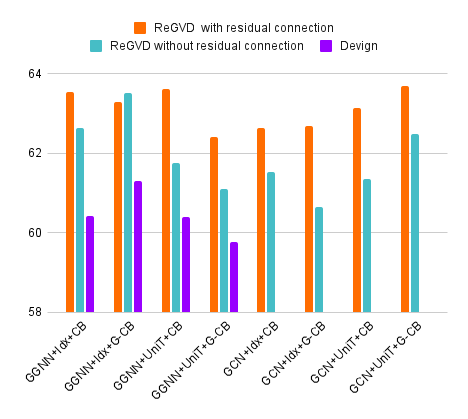}
  \caption{Accuracy with and without residual connection.}
  \label{fig:resconnection}
\end{subfigure}%
\begin{subfigure}{.245\textwidth}
  \centering
  \includegraphics[width=1\textwidth]{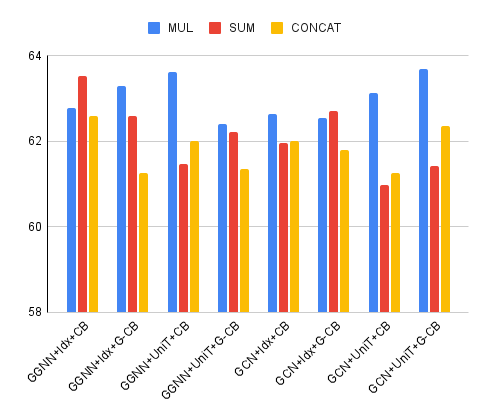}
  \caption{Accuracy w.r.t the $\mathsf{MIX}$ functions.}
  \label{fig:eff_mix_function}
\end{subfigure}
\caption{Accuracy with different settings.}
\label{fig:archi_e}
\end{figure}

We look at Figure \ref{fig:resconnection} 
to investigate whether the graph-level readout layer proposed in ReGVD performs better than the \textit{Conv} pooling layer utilized in Devign. 
Since Devign also uses Gated GNNs to update the node representations and gains the best accuracy of 61.31\% for the setting (Idx+G-CB); thus, we consider the ReGVD setting (GGNN+Idx +G-CB) without using the residual connection for a fair comparison, wherein ReGVD achieves an accuracy of 63.51\%, which is 2.20\% higher accuracy than that of Devign. 
More generally, we get a similar conclusion from the results of three remaining ReGVD settings (without using the residual connection) that the graph-level readout layer utilized in ReGVD outperforms that used in Devign.

We analyze the influence of the residual connection and the mixture function.
We first look back Figure \ref{fig:resconnection} for the ReGVD accuracies w.r.t with and without using the residual connection among the GNN layers.
It demonstrates that the residual connection helps to boost the GNNs performance on seven settings, where the maximum accuracy gain is 2.05\% for the ReGVD setting (GCN+Idx+G-CB).
Next, we look at Figure \ref{fig:eff_mix_function} for the ReGVD results w.r.t the $\mathsf{MIX}$ functions. We find that ReGVD generally gains the highest accuracies on six settings using the $\mathsf{MUL}$ operator and on two remaining settings using the $\mathsf{SUM}$ operator.
But it is worth noting that the ReGVD setting (GGNN+Idx+CB) using the $\mathsf{CONCAT}$ operator obtains an accuracy of 62.59\%, which is still higher than that of Devign, CodeBERT, and GraphCodeBERT.

Furthermore, our model achieves satisfactory performance with limited training data, compared to the baselines using the full training data.
For example, ReGVD obtains an accuracy of 61.68\% with 60\% training set, which is higher than the accuracies of BiLSTM, TextCNN, RoBERTa, and Devign.
It also achieves an accuracy of 62.55\% with 80\% training set, which is better than those of CodeBERT and GraphCodeBERT.




\section{Conclusion}  
\label{sec:conclusion}

We consider vulnerability identification as an inductive text classification problem and introduce a simple yet effective graph neural network-based model, named ReGVD, to detect vulnerabilities in source code. 
ReGVD transforms each raw source code into a graph, wherein ReGVD utilizes only the token embedding layer of the pre-trained programming language model to initialize node feature vectors.
ReGVD then leverages residual connection among GNN layers and a mixture of the sum and max poolings to learn graph representation.
To demonstrate the effectiveness of ReGVD, we conduct extensive experiments to compare ReGVD with the strong and up-to-date baselines on the benchmark vulnerability detection dataset from CodeXGLUE. 
Experimental results show that the proposed ReGVD is significantly better than the baseline models and obtains the highest accuracy of 63.69\% on the benchmark dataset.
ReGVD can be seen as a general, practical, and programming language-independent model that can run on various source codes and libraries without difficulty. 

\section*{Acknowledgements}
This research was partially supported under the Defence Science and Technology Group's Next Generation Technologies Program.
We would like to thank Anh Bui (tuananh.bui@monash.edu) for his help and support.

\bibliography{references}
\bibliographystyle{ACM-Reference-Format}
\end{document}